\pdfoutput=1

\documentclass[11pt]{article}

\usepackage[]{EMNLP2023}

\usepackage{times}
\usepackage{latexsym}

\usepackage[T1]{fontenc}

\usepackage[utf8]{inputenc}

\usepackage{microtype}

\usepackage{inconsolata}
\usepackage{url}
\usepackage{graphicx}
\usepackage{lscape}
\usepackage{amsmath}
\usepackage{multirow}
\usepackage{xcolor}

\title{Self-prompted Chain-of-Thought on Large Language Models for\\Open-domain Multi-hop Reasoning}

\author{Jinyuan Wang\textsuperscript{\rm 1, 3} \and Junlong Li\textsuperscript{\rm 2, 3} \and Hai Zhao\textsuperscript{\rm 2, 3}\thanks{\ \ Corresponding author. This paper was partially supported by the Joint Research Project of Yangtze River Delta Science and Technology Innovation Community (No. 2022CSJGG1400).} \\
        \textsuperscript{\rm 1}SJTU-Paris Elite Institute of Technology, Shanghai Jiao Tong University\\
	    \textsuperscript{\rm 2}Department of Computer Science and Engineering, Shanghai Jiao Tong University\\
	    \textsuperscript{\rm 3}Key Laboratory of Shanghai Education Commission for Intelligent Interaction\\
	    and Cognitive Engineering, Shanghai Jiao Tong University\\
	    \texttt{\{steve\_wang,lockonn\}@sjtu.edu.cn}, \texttt{zhaohai@cs.sjtu.edu.cn}
	    }

\begin{document}
\maketitle
\begin{abstract}
In open-domain question-answering (ODQA), most existing questions require single-hop reasoning  on commonsense. To further extend this task, we officially introduce open-domain multi-hop reasoning (ODMR) by answering multi-hop questions with explicit reasoning steps in open-domain setting. Recently, large language models (LLMs) have found significant utility in facilitating ODQA without external corpus. Furthermore, chain-of-thought (CoT) prompting boosts the reasoning capability of LLMs to a greater extent with manual or automated paradigms. However, existing automated methods lack of quality assurance, while manual approaches suffer from limited scalability and poor diversity, hindering the capabilities of LLMs. In this paper, we propose Self-prompted Chain-of-Thought (SP-CoT), an automated framework to mass-produce high quality CoTs of LLMs, by LLMs and for LLMs. SP-CoT introduces an automated generation pipeline of high quality ODMR datasets, an adaptive sampler for in-context CoT selection and self-prompted inference via in-context learning. Extensive experiments on four multi-hop question-answering benchmarks show that our proposed SP-CoT not only significantly surpasses the previous SOTA methods on large-scale (175B) LLMs, but also nearly doubles the zero-shot performance of small-scale (13B) LLMs. Further analysis reveals the remarkable capability of SP-CoT to elicit direct and concise intermediate reasoning steps by recalling $\sim$50\% of intermediate answers on MuSiQue-Ans dataset.
\end{abstract}

\section{Introduction}

Open-domain question-answering (ODQA) is a longstanding and challenging task which addresses factoid commonsense questions without specific contexts provided. While existing works in ODQA primarily focus on resolving questions that mostly require single-hop reasoning, there is a burgeoning interest in multi-hop question-answering (MHQA), which aims to derive the correct answer through multi-step reasoning over a collection of candidate articles \cite{mavi2022survey}. Yet, a significant disparity exists between such scenarios and real-world applications, since the latter often lacks an explicit set of candidate articles provided by users. In light of this, we officially introduce open-domain multi-hop reasoning (ODMR) as a progression task of ODQA, which requires MHQA with explicit rationales in open-domain setting. 

For ODMR, an emerging approach is to leverage large language models (LLMs) due to the vast knowledge stored within their numerous parameters. In recent years, LLMs have shown powerful reasoning and instruction-following capabilities, such as GPT-3 \cite{brown2020language}, PaLM \cite{chowdhery2022palm} and InstructGPT \cite{ouyang2022training}. After extensive training on vast corpora of textual data, LLMs prove to be zero-shot reasoners on complex reasoning tasks by breaking down multi-step questions into intermediate ones for step-by-step reasoning before producing the final answer \cite{kojima2023large}. Such series of intermediate reasoning steps is known as chain-of-thoughts (CoTs) \cite{wei2023chainofthought}. CoTs often serve as in-context demonstrations for in-context learning (ICL) \cite{brown2020language}, which enables LLMs to generate outputs that are formally consistent with a target task via a few reference examples provided as prompt. Manual-CoT \cite{wei2023chainofthought} adopt manually designed CoTs as in-context demonstrations to improve the reasoning performance of LLMs. However, it demands delicate and meticulous design by humans, and the demonstrations are the same for each question, which may be sub-optimal. Zero-shot-CoT \cite{kojima2023large} was proposed to trigger automated CoTs by certain specific prompting techniques, such as "\texttt{Let's think step by step:}". \citet{zhang2022automatic} proposed Auto-CoT, an automated framework to mass-produce CoTs and build in-context demonstrations. However, previous works have not fully leveraged the strong instruction-following and zero-shot reasoning capabilities of LLMs.

In this paper, we propose Self-prompted Chain-of-Thought (SP-CoT), an LLM-only framework to mass-produce high-quality CoTs for ODMR. In general, SP-CoT introduces an automated generation pipeline of ODMR datasets, an adaptive sampler for CoT selection and self-prompted inference via ICL. The automated ODMR datasets are MHQA datasets without candidate contexts, yet including multi-hop questions with six types of complex reasoning chains and step-by-step decomposition. Each intermediate QA step is equipped with a short explanation to justify the answer. By leveraging the ICL ability of LLMs, our method is generally effective on LLMs of different scales. 

We evaluate our method on four MHQA datasets in an open-domain setting: ComplexWebQuestions (CWebQ) \cite{talmor2018web}, HotpotQA \cite{yang-etal-2018-hotpotqa}, 2WikiMultiHopQA (2Wiki) \cite{ho-etal-2020-constructing} and MuSiQue-Ans (MSQ) \cite{trivedi2022musique}. Extensive experiments show that our proposed SP-CoT not only significantly surpasses the previous SOTA methods on large-scale (175B) LLMs, but also nearly doubles the zero-shot performance on small-scale (13B) LLMs in ODMR. Further analysis reveals the outstanding capability of SP-CoT to elicit direct and concise intermediate reasoning steps by recalling $\sim$50\% of intermediate answers on MSQ dataset.

Our contributions can be summarized as follows:

\begin{enumerate}
    \item We introduce an automated pipeline to generate high-quality ODMR datasets by LLMs, which include 2-4 hop questions with six types of complex reasoning chains.
    
    \item We propose SP-CoT, an automated framework to mass-produce CoTs while ensuring quality and diversity. 

    \item We conduct extensive experiments to confirm the effectiveness of SP-CoT on four ODMR benchmarks. In ODMR setting, our approach significantly boosts the performance by eliciting high-quality intermediate reasoning steps.
\end{enumerate}

Our code and datasets are publicly available at \href{https://github.com/noewangjy/SP-CoT}{https://github.com/noewangjy/SP-CoT}.

\begin{figure*}[htbp]
    \centering
    \includegraphics[scale=1.07]{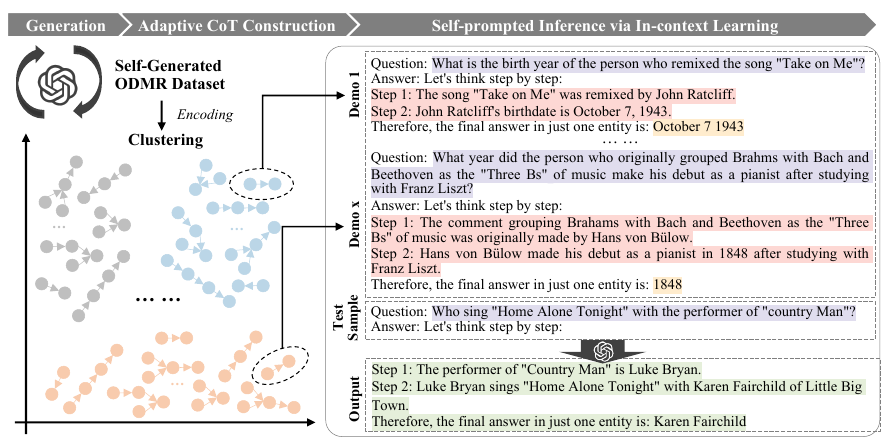}
    \caption{The overall framework of our proposed SP-CoT, including an automated generation of ODMR datasets, an adaptive sampler for CoT selection and self-prompted inference via ICL. Texts highlighted in purple refer to questions, in red to previously generated CoTs, in orange to answers, and in green to newly generated contents.}
    \label{fig:framework}
\end{figure*}

\section{Related Works}

\subsection{Multi-Hop Dataset Creation}

Creating an annotated MHQA dataset manually requires significant human resources. Therefore, some researchers  are dedicated to automating the generation of MHQA datasets. \citet{jiang2020hover} elaborated the creation of a multi-hop fact verification dataset from existing HotpotQA dataset. \citet{trivedi2022musique} introduced a bottom-up process to build challenging multi-hop reading comprehension QA dataset through meticulous selection and composition of single-hop questions derived from existing datasets. \citet{press2023measuring} proposed an automatically generated dataset with compositional 2-hop questions about celebrities. Nevertheless, existing approaches are either only partially automated, still requiring crowdsourcing, or they are limited to less complex 1-2 hop questions. In this work, our proposed SP-CoT is capable of automatically generating 2-4 hop questions with six different types of reasoning chains (Figure \ref{fig:mhqa} in Appendix).

\subsection{Chain-of-Thought Prompting}

Recent works on CoT prompting can be divided into two research lines. The first is prompting LLMs step by step to leverage their comprehension and reasoning abilities to answer questions. Zero-shot-CoT \cite{kojima2023large} adopts a two-stage design, which requires LLMs to first generate intermediate rationale and then produce an answer. \citet{wang2022iteratively} introduced iCAP, which iteratively prompts a fine-tuned small-scale LLM to generate CoTs and then combines the generated rationales to formulate answers. Least-to-Most \cite{zhou2023leasttomost} requires LLMs to first decompose a complex question into sub-questions and then sequentially solve them to arrive at the final answer. 

The second research line focuses on designing effective CoT as demonstrations for ICL to release more powerful reasoning abilities of LLMs. Manual-CoT \cite{wei2023chainofthought} was introduced to leverage manually designed CoT as in-context demonstrations to solve arithmetic, commonsense, and symbolic problems. A recent work \cite{zelikman2022star} shed light on the practicality to automate the generation of rationales by LLMs. Subsequently, Self-Ask \cite{press2023measuring} was proposed to construct step-by-step demonstrations with explicit decision process and intermediate answers as CoT. \citet{zhang2022automatic} proposed Auto-CoT, which automatically constructs CoTs via LLMs and adopts clustering methods to dynamically build demonstrations for each question.

However, existing methods have two significant limitations: 1) Over-reliance on the reasoning abilities of LLMs. Most methods are reported effective on large-scale LLMs like InstructGPT, while reproducing these methods on small-scale LLMs is quite challenging. 2) Over-confidence on the quality of intermediate results. When prompting LLMs step by step, defects in previous steps may limit the performance of subsequent steps. Similarly, while automatically constructing in-context demonstrations, the effectiveness of ICL might be compromised by the unstable quality of CoTs. Admittedly, manually constructed CoTs can ensure quality, yet they face a trade-off between content diversity and costs. To overcome the above drawbacks, our proposed SP-CoT automates CoT generation with quality ensured by leveraging the strong instruction-following capability of LLMs.

\begin{figure*}[htbp]
    \centering
    \includegraphics[scale=1.07]{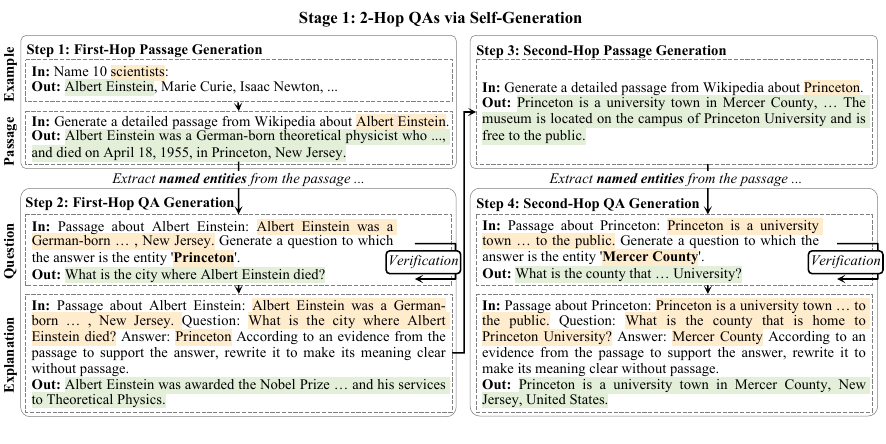}
    \caption{Generation steps for 2-hop QA quadruplets. Each QA quadruplet comprises a question $q$, its corresponding answer $a$, and a context passage $p$. The explanation $e$ includes the answer $a$ from the passage $p$ to address the question $q$. Text highlighted in orange refers to previously generated content, while the response of the LLM is highlighted in green.}
    \label{fig:stage1}
\end{figure*}

\subsection{Model Enhancement via LLM Generation}

With the powerful capabilities of LLMs on content generation and instruction-following, one recent research direction extensively leverages the content generated by LLMs to enhance smaller LLMs. Recent works such as GPTeacher,\footnote{\href{https://github.com/teknium1/GPTeacher}{https://github.com/teknium1/GPTeacher}} 
Alpaca \cite{alpaca} and Vicuna \cite{vicuna2023} collect the content generated by GPT-4 \cite{openai2023gpt4} and the corresponding prompts to train smaller-scale LLMs to achieve comparable performance. Another research line aims to boost the performance of large-scale LLMs to higher levels by leveraging the self-generated content. Some works use the self-generation as contexts to assist themselves in answering questions, such as eliciting intermediate rationales as CoT \cite{kojima2023large} or generating background articles for reading comprehension \cite{yu2023generate}. While others instruct LLMs to generate demonstrations for ICL during inference \cite{zhang2022automatic}, such as prompting LLMs to generate reliable QA pairs as self-prompted in-context demonstrations \cite{li2022self}. Our work is dedicated to extending the second research line to ODMR by leveraging the automated self-generated CoT as in-context demonstrations. Compared to previous works \cite{kojima2023large, zhang2022automatic, li2022self}, our work taps into the potential of self-prompting LLMs with more complicated framework design to solve a most challenging task.

\section{Methods}
\label{sec:method}

In this section, we elaborate our proposed SP-CoT in three stages (Figure \ref{fig:framework}):

In the first stage, we prompt LLMs to iteratively generate 2-hop commonsense QA quadruplets with context, question, answer and explanation. 

In stage 2, we construct multi-hop reasoning chains by connecting 2-hop QA quadruplets and build an ODMR dataset via composition. 

In the last stage, we adopt clustering-based sampling approach to dynamically select and construct in-context demonstrations for inference.

\subsection{2-Hop QAs via Self-Generation}

In the first stage, we prompt LLMs to iteratively generate 2-hop QA quadruplets with context, question, answer and explanation, which is illustrated in Figure \ref{fig:stage1}. Inspired by \citet{li2022self}, we design a 2-hop commonsense QA generation pipeline, including the following 4 steps:

\textbf{Step 1: First-Hop Passage Generation}
To guarantee the comprehensive coverage of commonsense knowledge, we manually design 29 diverse topics based on the statistics of TriviaQA \cite{joshi-etal-2017-triviaqa}. For each topic, we require the LLM to name a certain number of keywords. For each collected keyword $k_1$, we ask the LLM to generate a Wiki-style passage $p_1$. Despite some factoid errors \cite{li2022self}, such generated passages contain sufficient factual information to serve as context for QA generation.

\textbf{Step 2: First-Hop QA Generation}
Given that the answers for commonsense questions are likely to be named entities, we use Spacy\footnote{\href{https://spacy.io/}{https://spacy.io/}} and NLTK \cite{bird-loper-2004-nltk} libraries to extract the named entities in the passage $p_1$ as candidate answers. For each candidate answer $a_1$, we require the LLM to raise a question $q_1$ to which the answer is $a_1$ based on the passage $p_1$. To ensure the quality of $q_1$, we employ a double-check process, where we demand the LLM to answer the generated question $q_1$ given the context $p_1$ to check if the generated answer $a_1$\textsuperscript{\rm $'$} is accordant with $a_1$. Once the generated QA pair passes the double-check, we prompt the LLM to write a short explanation $e_1$ for it. Note that the candidate answers must exclude the keyword ($a_1\neq k_1$) because the answer in the first hop will become the keyword for the second hop ($k_2 = a_1$, $k_2 \neq k_1$). In addition to that, a valid explanation must contain the answer ($a_1\in e_1$).

\begin{figure*}[htbp]
    \centering
    \includegraphics[scale=1.07]{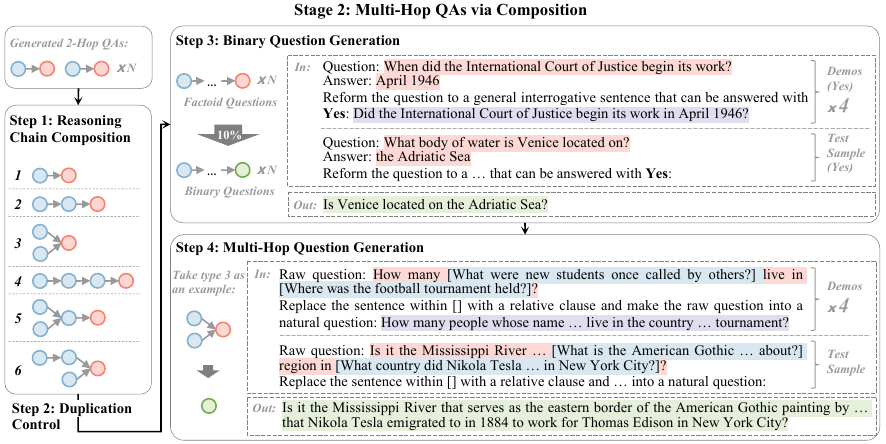}
    \caption{Generation steps for MHQA groups. In step 3 and step 4, we use 4 manually designed demonstrations for ICL. Each MHQA group includes a multi-hop question, the corresponding answer and decomposed QA quadruplets. Nodes and texts highlighted in red, blue and green successively refer to the last hops, intermediate hops and generated hops. Manually designed texts are highlighted in purple.}
    \label{fig:stage2}
\end{figure*}

\textbf{Step 3: Second-Hop Passage Generation}
Before the first-hop answers are used as keywords for second-hop passage generation, we use Spacy to filter out the answers with certain labels (\texttt{QUANTITY}, \texttt{ORDINAL}, \texttt{CARDINAL}, \texttt{PERCENT}, \texttt{MONEY}, \texttt{DATE}, \texttt{TIME}), which are infeasible for Wiki-style passage generation. Given a keyword $k_2$, we repeat the same prompts as described in Step 1 to generate the passage $p_2$.

\textbf{Step 4: Second-Hop QA Generation}
We begin with extracting candidate answers in the generated passage $p_2$ while blocking the keyword $k_1$ and the answer $a_1$ (also known as $k_2$) in the first-hop QA to avoid cyclic graphs. For each candidate answer $a_2$, we require the LLM to generate a question $q_2$ which contains the first-hop answer $a_1$ and can be answered by the candidate answer $a_2$. We examine the quality of $q_2$ with the same double-check in Step 2, and ensure the second-hop question $q_2$ contains the first-hop answer $a_1$ ($a_1 \in q_2$) for connected reasoning. Then we repeat the same prompts in Step 2 to generate explanation $e_2$.

So far, we have instructed the LLM to generate a 2-hop commonsense QA quadruplet pair, which is $(p_1, q_1, a_1, e_1) \to (p_2, q_2, a_2, e_2)$ with $a_1 \in q_2$. Detailed prompt templates are shown in Figure \ref{fig:stage1} and Appendix \ref{app:prompt}.

\begin{table*}[htbp]
\centering
\caption{Comparison of different approaches on four MHQA benchmarks. The fine-tuning methods are fine-tuned on the train split of NQ \cite{kwiatkowski-etal-2019-natural} dataset. Among them, methods marked with "\textsuperscript{\rm $\dagger$}" use the Wikipedia dump \cite{karpukhin-etal-2020-dense} as extra corpus. For retrieval-based methods, we use a fine-tuned DPR \cite{karpukhin-etal-2020-dense} to retrieve top-5 documents from Wikipedia as context and employ LLM as Reader to answer the question based on the context. Methods based on ChatGPT are performed by \texttt{gpt-3.5-turbo-0301} version.} 
\label{tab:result}
\small
\begin{tabular}{lcccccccccc}
\hline
\multirow{2}{*}{\textbf{Methods}} &
  \multicolumn{2}{c}{\textbf{MSQ}} &
  \multicolumn{2}{c}{\textbf{HotpotQA}} &
  \multicolumn{2}{c}{\textbf{2Wiki}} &
  \multicolumn{2}{c}{\textbf{CWebQ}} &
  \multicolumn{2}{c}{\textbf{Average}} \\
 &
  \textbf{EM} &
  \textbf{F1} &
  \textbf{EM} &
  \textbf{F1} &
  \textbf{EM} &
  \textbf{F1} &
  \textbf{EM} &
  \textbf{F1} &
  \textbf{EM} &
  \textbf{F1} \\ \hline

\textbf{\textit{Fine-tuning methods with extra corpus}} \\

DPR\textsuperscript{\rm $\dagger$} \text{\cite{karpukhin-etal-2020-dense}}        & 7.5  & 16.6       & 14.7 & 23.0       & 6.5  & 14.6        & 21.3 & 30.2      & 12.5 & 21.1 \\ 
RAG\textsuperscript{\rm $\dagger$} \text{\cite{lewis2020retrieval}}               & 3.9  & 8.2        & 9.6  & 15.5       & 15.7 & 18.6        & 13.3 & 15.8      & 10.6 & 14.5 \\ 
REALM\textsuperscript{\rm $\dagger$}  \text{\cite{guu2020realm}}                  & 3.7  & 8.5        & 15.3 & 22.0       & 5.1  & 9.6         & 21.1 & 27.3      & 11.3 & 16.9 \\ 
T5-11B-SSM  \text{\cite{roberts-etal-2020-much}}            & 10.6 & 16.6       & 15.4 & 22.4       & 15.6 & 20.5        & 28.5 & 35.5      & 17.5 & 23.8 \\ \hline

\textbf{\textit{Retrieval-based methods with LLMs}} \\

DPR\textsuperscript{\rm $\dagger$}  + ChatGPT                                                       & 1.7  & 4.1        & 15.8 & 21.4       & 10.9 & 18.1       & 13.7 & 18.7       & 10.5 & 15.6 \\ 
DPR\textsuperscript{\rm $\dagger$} + Alpaca-13B \cite{alpaca}                                      & 2.2  & 8.4        & 12.0 & 21.5       & 12.6 & 21.2       & 15.9 & 27.1       & 10.7 & 19.5 \\
DPR\textsuperscript{\rm $\dagger$} + Vicuna-13B \cite{vicuna2023}                                  & 2.2  & 7.4        & 18.6 & 25.8       & 23.9 & 27.6       & 20.3 & 27.9       & 16.2 & 22.1 \\
DPR\textsuperscript{\rm $\dagger$} + WizardLM-13B \cite{xu2023wizardlm}                            & 3.5  & 10.0       & 19.9 & 28.4       &  22.8 & 27.5       & 24.2 & 32.2       & 17.6 & 24.5 \\
DPR\textsuperscript{\rm $\dagger$} + InstructGPT \cite{ouyang2022training}                         & 4.8  & 11.6       & 26.3 & 34.8       & 23.3 & 27.1       & 34.4 & 41.6       & 22.2 & 28.8 \\

\hline

\textbf{\textit{LLM-only methods on ChatGPT}} \\

Zero-Shot                                                           & 3.1  & 7.3        & 22.4 & 30.0       & 18.7 & 21.7       & 31.6 & 37.5       & 19.0 & 24.1 \\
Self-Prompting      \text{\cite{li2022self}}                        & 2.9  & 6.2        & 23.8 & 31.2       & 18.9 & 23.5       & 26.8 & 32.6       & 18.1 & 23.4 \\
GENREAD      \text{\cite{yu2023generate}}                           & 8.6  & 14.6       & \textbf{33.2} & 42.6                  &\textbf{30.4} & \textbf{35.3}          & 33.7 & 40.1           & 26.5 & 33.2 \\ 
Zero-shot-CoT       \text{\cite{kojima2023large}}                   & 5.0  & 8.8        & 22.6 & 29.6       & 24.3 & 27.1       & 30.3 & 36.2       & 20.6 & 25.4 \\ 
Auto-CoT        \text{\cite{zhang2022automatic}}                    & 8.1  & 13.6       & 26.1 & 36.3       & 26.2 & 30.2        & 29.9 & 38.4      & 22.6 & 29.6 \\ 
Manual-CoT (random) \text{\cite{wei2023chainofthought}}             & 12.3  & 19.2       & 32.4 & \textbf{43.7}      & 27.7 & 34.6        & 36.6 & 43.0      & 27.3 & 35.1 \\ 
SP-CoT \textit{(Ours)} &
  \textbf{14.5} &
  \textbf{22.6} &
  \textbf{33.2} &
  42.9 &
  30.1 &
  34.7 &
  \textbf{37.5} &
  \textbf{43.6} &
  \textbf{28.8} &
  \textbf{36.0} \\ \hline
\end{tabular}
\end{table*}

\subsection{Multi-Hop QAs via Composition}

In stage 2, we construct multi-hop reasoning chains with the connected 2-hop QA quadruplets, which is illustrated in Figure \ref{fig:stage2}. We propose an automated dataset construction pipeline to build ODMR datasets with 2-4 hops, which has the following 4 steps:

\textbf{Step 1: Reasoning Chain Composition}
To connect more questions, we follow the composability criteria \cite{trivedi2022musique}, that is, two single-hop QA pairs $(q_1, a_1)$ and $(q_2, a_2)$ are composable into a multi-hop question $Q$ with $a_2$ as a valid answer if $a_1$ is a named entity and it is mentioned in $q_2$. Such criteria are already satisfied when our 2-hop QA pairs are generated, we use this criterion for connecting more questions. We adopt 6 reasoning graphs with 2-4 hops to build 6 types of multi-hop reasoning chains (Figure \ref{fig:mhqa} in Appendix), and we ensure that in each reasoning chain: 1) the answer $a_i$ to an intermediate question $q_i$ will appear and ONLY appear in its next-hop question $q_{i+1}$ to avoid shortcuts; 2) the answer to the last question will NOT appear in any intermediate questions.

\textbf{Step 2: Duplication Control}
Built by rule-based composition, our new dataset has considerably similar reasoning chains that have duplicate intermediate questions. To ensure the diversity and simplicity of our dataset, we filter out the reasoning chains by a preset duplication degree which is defined by the number of questions that co-existed in other chains within the same reasoning type.

\textbf{Step 3: Binary Question Generation}
We notice that MHQA datasets also include general interrogative questions which should be answered by "Yes" or "No", rather than a named entity. Therefore, we leverage the LLM to reform the last QA $(q_n, a_n)$ of some reasoning chains to binary question with 4 manually designed in-context demonstrations. For each reasoning type, we randomly sample 10\% reasoning chains for positive question generation and 10\% for negative ones. Then we reform a new reasoning chain by the generated binary question together with its previous question hops and add it to the dataset.

\textbf{Step 4: Multi-Hop Question Generation}
Now we need to generate multi-hop questions, to which the previously generated question chains serve as their intermediate reasoning steps. For each question chain, we iteratively replace the answer $a_i$ to an intermediate question $q_i$ in the next-hop question $q_{i+1}$ by $[q_i]$ until the last question $q_n$ is replaced, which indicates a relative clause. Then we leverage the LLM to reform it into a natural multi-hop question with 4 manually designed in-context demonstrations.

After the pipeline above, we construct a high quality ODMR dataset with 2-4 hops, including the overall multi-hop question, the decomposed reasoning chains with detailed QA quadruplets. With the double-check in generation and the composability criteria, we automatically build a high quality new dataset. Detailed prompt templates are presented in Figure \ref{fig:stage2} and Appendix \ref{app:prompt}.

\subsection{Adaptive In-context Demonstration}

In this stage, we sample multi-hop questions from our generated ODMR dataset as in-context demonstrations.

\textbf{Clustering-based Retrieval}
Some previous works \cite{zhang2022automatic, li2022self} have shown that clustering-based methods benefit from the diversity of demonstrations. We adopt a clustering-based retrieval approach to adaptively sample in-context demonstrations for the input question. First, all the questions are projected to a high dimension hidden space by encoding with Sentence-BERT \cite{reimers-gurevych-2019-sentence}. Suppose we need $n$ in-context demonstrations. Given a test question $Q$, we use \textit{k-means} to cluster the question embeddings into $n$ clusters and adaptively retrieve the question with the highest cosine similarity to $Q$ from each cluster.

\textbf{Build Reasoning Chain}
For each sampled example, we sequentially concatenate the explanation from each hop,  prefaced by "\texttt{Step \{$i$\}:}", to construct a reasoning chain.

\begin{table*}[htbp]
\centering
\caption{The performance (EM) of our method on recent popular LLMs. We use \texttt{text-davinci-003} for InstructGPT. On small-scale LLMs, SP-CoT nearly doubles the average zero-shot performance on four MHQA benchmarks.}
\label{tab:general_test}
\small
\begin{tabular}{lcccccccc}
\hline
\textbf{Model} & \textbf{Size} & \textbf{Method} & \textbf{MSQ} & \textbf{HotpotQA} & \textbf{2Wiki} & \textbf{CWebQ} & \textbf{Mean} & \textbf{Boost} \\ \hline
Alpaca \cite{alpaca} & 13B       & Zero-shot          & 0.9  & 9.5  & 12.5 & 12.6 & 8.9 & - \\
Alpaca \cite{alpaca} & 13B       & SP-CoT & 5.3  & 19.8 & 17.8 & 26.5 & 17.4 & 8.5$\uparrow$ \\ \hline
Vicuna \cite{vicuna2023} & 13B       & Zero-shot          & 2.3  & 11.5 & 18.1 & 18.3 & 12.6 & - \\
Vicuna \cite{vicuna2023} & 13B       & SP-CoT & 8.5  & 23.3 & 22.0 & 32.2 & 21.5 & 8.9$\uparrow$ \\ \hline
WizardLM \cite{xu2023wizardlm} & 13B       & Zero-shot          & 2.0  & 10.3 & 13.6 & 17.5 & 10.9 & - \\
WizardLM \cite{xu2023wizardlm} & 13B       & SP-CoT & 7.3  & 24.0 & 27.2 & 31.3 & 22.5 & 11.6$\uparrow$ \\ \hline
InstructGPT \cite{ouyang2022training} & 175B & Zero-shot          & 4.4  & 25.6 & 22.8 & 38.6 & 22.9 & - \\
InstructGPT \cite{ouyang2022training} & 175B & SP-CoT & 14.8 & 37.4 & 33.7 & 46.6 & 33.1 & 10.2$\uparrow$ \\ \hline
\end{tabular}

\end{table*}

\section{Experiments}

Our research questions (RQs) are:

RQ1: \textit{To what extent can SP-CoT boost the LLMs on our four ODMR benchmarks, compared with other LLM-only methods?}

RQ2: \textit{Is SP-CoT generally effective on recent popular instruction-following LLMs?}

To this end, we conduct experiments on four MHQA datasets that require complex multi-step reasoning and compare different methods across different LLMs.

\subsection{Benchmarks and Evaluation Metrics}

We choose the following four MHQA datasets: CWebQ, HotpotQA, 2Wiki and MSQ. We set them as ODMR benchmarks by taking only the question and the answer in each example. Dataset introduction and statistics are detailed in Appendix \ref{app:dataset}.

We adopt the exact match (EM) and F1 scores as our evaluation metrics. Based on the evaluation script of \citet{karpukhin-etal-2020-dense}, we add a preprocessing step which ignores the content within "\texttt{()}" and splits the answer strings by certain delimiters to extract multiple answers.

\subsection{Experiment Settings}

For reference, we experiment with fine-tuning methods using an extra corpus, which are fine-tuned on the training split of NQ \cite{kwiatkowski-etal-2019-natural} dataset and most of them adopt the Wikipedia dump \cite{karpukhin-etal-2020-dense} as extra corpus. We also test our implementation of the retrieval-based methods on most recent LLMs for reference. Specifically, we use a fine-tuned DPR \cite{karpukhin-etal-2020-dense} to retrieve top-5 documents from Wikipedia as context and employ LLM as Reader to answer the question based on the context. Detailed prompt templates and parameter settings are provided in the Appendix \ref{app:prompt}.

Unless otherwise specified, we use Sentence-BERT (\texttt{all-mpnet-base-v2}) for question encoding following previous works. The default number of in-context demonstrations is 8 and the demonstrations are sampled by the maximum cosine similarity of questions in each cluster. 

For RQ1, we adopt ChatGPT (\texttt{gpt-3.5-turbo- 0301}) as the LLM to conduct the following experiments. According to OpenAI,\footnote{\href{https://platform.openai.com/docs/model-index-for-researchers}{https://platform.openai.com}} \texttt{gpt-3.5-turbo- 0301} is an improvement on the InstructGPT \texttt{text- davinci-003} model, which performs at a similar capability level to \texttt{text-davinci-003} for inference. We use the whole development set of each dataset in our experiments. 

For RQ2, we not only test InstructGPT (\texttt{text- davinci-003}), but also employ three smaller-scale (13B) LLMs: Alpaca \cite{alpaca}, Vicuna \cite{vicuna2023} and WizardLM \cite{xu2023wizardlm}, which are LLaMA \cite{touvron2023llama} models fine-tuned on different large-scale instruction-following datasets. To save computational cost, we conduct this experiment on subsets of the four datasets by randomly selecting 1000 samples from the test sets.

\begin{table*}[htbp]
\centering
\small
\caption{The performance (EM) of different methods of demonstration selection. The results from random selection represent the mean value and standard deviation obtained from 3 runs, each with a different seed.}
\label{tab:selection}
\begin{tabular}{lccccc}
\hline
\textbf{Method}                                                    & \textbf{MSQ} & \textbf{HotpotQA} & \textbf{2Wiki} & \textbf{CWebQ} & \textbf{Average} \\ \hline
Random                                                             & \textbf{12.0}$_{\pm0.8}$    & 29.5$_{\pm0.6}$             & 25.6$_{\pm1.2}$           & 34.3$_{\pm0.9}$           & 25.4$_{\pm0.3}$             \\ 
Retrieve                                                           & 10.5             & 27.9              & 24.3           & 33.5           & 24.1             \\ 
ClusterCenter                                                     & 10.4             & 26.2              & 22.8           & 33.0           & 23.1             \\ 
RetrieveInTypeCluster & 11.6             & 28.7              & 23.5           & \textbf{35.3}  & 24.8             \\ 
RetrieveInCluster & 11.5 & \textbf{30.9} & \textbf{27.8} & 34.0 & \textbf{26.1} \\ \hline
\end{tabular}
\end{table*}

\subsection{Experiment Results}

The main results of RQ1 are shown in Table \ref{tab:result}. Even with extra corpus, the models fine-tuned on NQ \cite{kwiatkowski-etal-2019-natural} present poor performance due to the inherent challenges of MHQA. With the same Retriever model, the performance of retrieval-based methods depends largely on the LLM Readers. Compared to previous LLM-only works, our SP-CoT significantly outperforms the Auto-CoT by $+6.2$ EM and $+6.4$ F1 scores on average and surpasses the previous SOTA method GENREAD \cite{yu2023generate} by $+2.3$ EM and $+2.8$ F1 scores on average. On the most challenging benchmark MSQ, SP-CoT empowers ChatGPT to outperform other LLM-only methods by a decent margin.

\begin{figure}[htbp]
    \centering
    \includegraphics[scale=0.9]{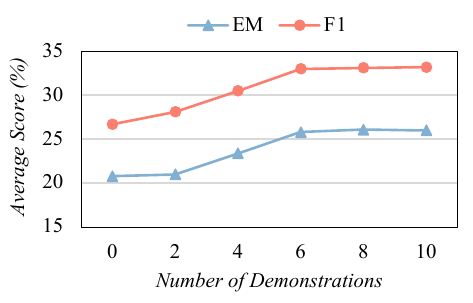}
    \caption{Average EM and F1 scores of different numbers of in-context demonstrations. The experiments are tested on 1k subsets of four ODMR benchmarks with ChatGPT (\texttt{gpt-3.5-turbo-0301}).}
    \label{fig:demo}
\end{figure}

We notice that SP-CoT significantly outperforms GENREAD on MSQ, confirming the effectiveness of providing high quality CoTs as in-context demonstrations for complex multi-hop questions. On the other three datasets, SP-CoT delivers comparable performance with GENREAD. However, GENREAD relies heavily on the generation faithfulness of LLMs, which is challenging for small-scale LLMs. By breaking down demanding instructions into step-by-step simple ones, our method is more applicable to small-scale LLMs, which is validated by Table \ref{tab:general_test}.

Table \ref{tab:general_test} presents the results for RQ2. Our proposed SP-CoT proves to be generally effective by significantly boosting the performance of all these four LLMs on all four benchmarks. With SP-CoT, the performance of small-scale (13B) LLMs can be boosted to be on par with directly prompting LLMs that are over 10$\times$ larger, regardless of the elicited high quality intermediate reasoning steps.

\section{Analysis}

In this section, we explore the choices of the sampling methods and the number of demonstrations. Then we examine the quality of the intermediate reasoning steps elicited by SP-CoT and the quality of self-generation data. Unless otherwise specified, we use ChatGPT (\texttt{gpt-3.5-turbo-0301}) to conduct analysis on the same subsets mentioned in RQ2 settings.

\begin{figure}[htbp]
    \centering
    \includegraphics[scale=0.9]{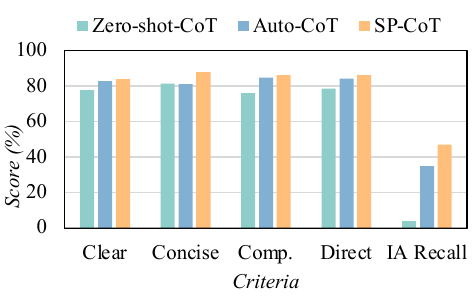}
    \caption{Evaluation results of the CoT generated by three methods. The first four scores are in terms of clearness, conciseness, comprehensibility (Comp.) and directness given by GPT-4 on 50 examples. The recall accuracy of intermediate answers (IA Recall) is reported on the questions that are correctly answered by all 3 methods.}
    \label{fig:eval}
\end{figure}

\subsection{Methods of Demonstration Sampling}

The performance of ICL depends largely on the quality of demonstration sampling. We test the effectiveness of the following five strategies: randomly sampling (Random), sampling globally by maximum cosine similarity (Retrieve), sampling the closest to the centroid in each cluster (ClusterCenter), sampling by the maximum cosine similarity in each cluster (RetrieveInCluster) and sampling the most similar QAs in each cluster in a certain reasoning type (RetrieveInTypeCluster). The reasoning type of the input question is determined by the most frequent reasoning type of its k-nearest neighbors. As indicated in Table \ref{tab:selection}, RetrieveInCluster \cite{li2022self} is the best-performing strategy, which is exactly the strategy we adopt in previous experiments.

\subsection{Impact of Demonstration Amount}

Providing more in-context demonstrations empirically improves ICL performance; however, it also causes increasing computational cost. To this end, we investigate the trade-offs of number of demonstrations and the resulting performance boost. We report the EM and F1 scores over the four benchmarks for 2, 4, 6, 8, and 10 in-context demonstrations, as well as the scores in a zero-shot setting. As illustrated in Figure \ref{fig:demo}, the performance of SP-CoT increases with the number of demonstrations when the count is between 2 and 8; however, using 10 demonstrations doesn't yield any further performance boost. In our main experiments, we opted for 8 as the default number of demonstrations, striking a balance between performance and cost.

\subsection{Intermediate Reasoning Quality Analysis}

Given the high-quality CoTs constructed by our proposed SP-CoT, we investigate the quality of intermediate reasoning steps generated during inference. For this analysis, we use the development set of MSQ, as it's the most challenging of the four datasets and offers decomposed step-by-step QAs. We compare the CoTs generated by Zero-shot-CoT, Auto-CoT and SP-CoT during inference. For fairness, we select 50 out of a total of 59 questions that all of the three methods answered correctly. First, we use GPT-4 to evaluate\footnote{Script from \href{https://github.com/lm-sys/FastChat/tree/main/fastchat/eval}{https://github.com/lm-sys/FastChat} and modified.} the intermediate reasoning steps in terms of clearness, conciseness, comprehensibility and directness separately on a scale of 1 to 10. Additionally, we compute the recall accuracy of intermediate answers co-occurring in the reasoning steps of each method. For fairness, we only report the intermediate answer recall accuracy of correctly answered questions for each method. As depicted in Figure \ref{fig:eval}, GPT-4 highly favors our SP-CoT, which achieves nearly a 50\% recall accuracy for intermediate answers. This suggests that SP-CoT elicits high-quality reasoning steps in terms of clearness, conciseness, comprehensibility, and directness.

\section{Conclusion}

In this work, we harness the capabilities of LLMs combined with self-prompted CoTs to tackle the intricate MHQA task within the open-domain context, termed as ODMR. Our innovative SP-CoT not only sets a new benchmark by surpassing preceding CoT prompting techniques but also outclasses the erstwhile SOTA LLM-only methodologies in open-domain question-answering. A distinguishing feature of SP-CoT is its proficiency in eliciting high-caliber intermediate reasoning steps, and its universal efficacy across both large and small-scale LLMs. We anticipate our innovative self-generation pipeline for ODMR to not just be foundational for SP-CoT, but also to pave the way for future research, catalyzing a shift towards leveraging self-generation in LLMs, by LLMs, and for LLMs.

\section*{Acknowledgements}

This paper was partially supported by the Joint Research Project of Yangtze River Delta Science and Technology Innovation Community (No. 2022CSJGG1400) and under the technical support of National Key R\&D Program of China (No. 2021YFC3340700). Our appreciation also extends to Ms. Yangyang Ding, who has been generous with her time and knowledge, offering constructive criticism and enriching discussions.

\section*{Limitations}
Our proposed method (SP-CoT) leverages the strong instruction-following power of LLMs. Such capability is easy to acquire through instruction fine-tuning for small-scale LLMs (even 7B), however, some LLMs proposed in early years may show poor capability in following human instructions due to lack of corresponding training before their release. Therefore, the performance of such LLMs may not be boosted by our proposed SP-CoT. In fact, we did not succeed in boosting the performance of GPT-NeoX by any of Zero-shot-CoT, Auto-CoT and SP-CoT. GPT-NeoX is a 20B LLMs released in early 2022, which shows poor instruction-following capability. Please note that neither GENREAD \cite{yu2023generate} nor Self-prompting \cite{li2022self}  boosts the performance of GPT-NeoX.

It is acknowledged that the efficacy of LLM-only approaches is predominantly reliant on the LLMs themselves. With smaller-scale LLMs, specifically those of 13B scale, our SP-CoT together with other CoT methodologies, demonstrate comparable or similar performance enhancement across four ODMR benchmarks as presented in Table \ref{tab:general_test2}. The consistent performance of handcrafted CoTs remains ambivalent across different LLMs and benchmarks; our empirical observations indicate that Manual-CoT occasionally outperforms SP-CoT, while at other instances, it does not.

Given the potential for LLMs to generate imprecise information, the process by which our SP-CoT produces datasets might also result in the emergence of inaccurate QA pairs as well as erroneous explanations. Despite the incorporation of a double-check mechanism to ensure data integrity, certain errors and inaccuracies are inevitably present.

\bibliography{anthology,custom}
\bibliographystyle{acl_natbib}

\appendix

\section{Datasets}
\label{app:dataset}

\subsection{Introduction}
\textbf{HotpotQA \cite{yang-etal-2018-hotpotqa}} HotpotQA is a widely used dataset for multi-hop question-answering (MHQA), which contains 113k multi-hop questions in natural language. The questions are collected by crowdsourcing based on Wikipedia articles with human annotated supporting evidence and answers.

\textbf{2WikiMultiHopQA \cite{ho-etal-2020-constructing}} 2WikiMultiHopQA s a recently proposed large-scale MHQA dataset, which contains over 192k samples constructed jointly from Wikipedia and Wikidata.

\textbf{MuSiQue-Ans \cite{trivedi2022musique}} MuSiQue-Ans (MSQ) is a recent challenging MHQA dataset created via single-hop question composition. It includes 25k 2-4 hop questions with six different composition structures. Although MSQ is composed from existing datasets, it poses 3$\times$ the human-machine gap with a substantially lower disconnected reasoning score.

\textbf{ComplexWebQuestions \cite{talmor2018web}} ComplexWebQuestions is a manually generated MHQA dataset of 35k QA pairs. CWebQ is generated by rephrasing questions generated by machine from existing dataset.

\subsection{Statistics}

The statistics of four datasets are shown in Table \ref{tab:dataset}.

\begin{table}[h]
\centering
\caption{Statistics of development set of four MHQA benchmarks, average tokens for questions, answers and average number of reasoning steps. CWebQ does not provide supporting evidence or question decomposition.}
\label{tab:dataset}
\small
\begin{tabular}{lcccc}
\hline
 & \textbf{MSQ} & \textbf{HotpotQA} & \textbf{CWebQ} & \textbf{2Wiki} \\
\hline
Q len. & 18.11 & 15.83 & 13.37 & 11.98 \\
A len.  & 2.8   & 2.46  & 2.42  & 2.41  \\
Steps  & 2.65  & 2.68  & -     & 2.47  \\
Size & 2417 & 7405 & 3519 & 12576 \\
\hline
\end{tabular}
\end{table}

\section{Prompt Templates}
\label{app:prompt}

\subsection{First-Hop QA Generation}

We following the notations described in Section \ref{sec:method}. The templates are:

\begin{enumerate}
    \item Name \{Number\} \{Topic\}:
    \item Generate a Wikipedia passage about \{$k_1$\}.
    \item Passage about \{$k_1$\}:\textbackslash n\{$p_1$\}\textbackslash n\textbackslash nGenerate a question to which the answer is the entity \{$a_1$\}. 
    \item Passage about \{$k_1$\}:\textbackslash n\{$p_1$\}\textbackslash n\textbackslash nQuestion:\textbackslash n \{$q_1$\}\textbackslash n\textbackslash nExtract the answer directly from the passage in less words as possible.
    \item Passage about \{$k_1$\}:\textbackslash n\{$p_1$\}\textbackslash n\textbackslash n Question:\textbackslash n \{$q_1$\}\textbackslash n\textbackslash nAnswer:\textbackslash n\{$a_1$\}\textbackslash n\textbackslash nAccording to an evidence from the passage to support the answer, rewrite it to make its meaning clear without passage.
\end{enumerate}

\subsection{Second-Hop QA Generation}

\begin{enumerate}
    \item Generate a Wikipedia passage about \{$k_2$\}.
    \item Passage about \{$k_2$\}:\textbackslash n\{$p_2$\}\textbackslash n\textbackslash nGenerate a question that meets the following conditions: 1. contains the term '\{$k_2$\}' in question, 2. the answer is \{$a_2$\}, 3. Avoid the following entities in the question: \{$k_2$\}
    \item Passage about \{$k_2$\}:\textbackslash n\{$p_2$\}\textbackslash n\textbackslash nQuestion:\textbackslash n \{$q_2$\}\textbackslash n\textbackslash nExtract the answer directly from the passage in less words as possible.
    \item Passage about \{$k_2$\}:\textbackslash n\{$p_2$\}\textbackslash n\textbackslash n Question:\textbackslash n \{$q_1$\}\textbackslash n\textbackslash nAnswer:\textbackslash n\{$a_2$\}\textbackslash n\textbackslash nAccording to an evidence from the passage to support the answer, rewrite it to make its meaning clear without passage.
\end{enumerate}

\subsection{Binary Question Generation}

\begin{itemize}
    \item Question: \{$q_n$\}\textbackslash nAnswer: \{$a_n$\}\textbackslash nReform the question to a general interrogative question that can be answered with yes:
    \item Question: \{$q_n$\}\textbackslash nAnswer: \{$a_n$\}\textbackslash nReform the question to a general interrogative question that can be answered with no:
\end{itemize}

\subsection{Multi-Hop Question Generation}

\begin{itemize}
    \item Raw question: \{$q_n$\}\textbackslash nReplace the sentence within [] with a relative clause and make the raw question into a natural question:
\end{itemize}

\subsection{CoT Construction}

Suppose $q^*$ is the generated multi-hop question, $e_i$ denotes the explanation from intermediate hop ($q_i$, $a_i$, $e_i$), $a^*$ is the answer of the last hop ($a^* = a_n$). The template is:

\begin{itemize}
    \item Question: \{$q^*$\}\textbackslash nAnswer: Let's think step by step:\textbackslash nStep 1: \{$e_1$\}\textbackslash nStep 2: \{$e_2$\}\textbackslash n ... Therefore, the answer in just one entity is: \{$a^*$\},
\end{itemize}

\subsection{Inference}

\subsubsection{Zero-shot}

Given a question $Q$, the inference template is:

\begin{itemize}
    \item Answer the following question with just one entity:\textbackslash nQuestion: \{$Q$\}\textbackslash nAnswer: 
\end{itemize}

\subsubsection{SP-CoT}
Suppose we have the input question $Q$ and 2 demonstrations ($q_1$, $r_1$, $a_1$), ($q_2$, $r_2$, $a_2$), where $q_i$, $r_i$, $a_i$ denote the question, CoT and answer of the $i^{th}$ demonstration. The inference template is:

\begin{itemize}
    \item Question: \{$q_1$\}\textbackslash n\{$r_1$\}\textbackslash n\textbackslash nQuestion: \{$q_2$\}\textbackslash n \{$r_2$\}\textbackslash n\textbackslash nQuestion: \{$Q$\}\textbackslash nAnswer: Let's think step by step:\textbackslash n
\end{itemize}

\begin{table*}[htbp]
\centering
\small
\caption{Performance (EM/F1) of additional CoT variants. In our experiment, Manual-CoT (Cherry-Pick) adopts 8 cherry-picked questions and their CoTs manually writen by the authors. The results of Manual-CoT (Random) report the mean EM scores of randomly selected questions and theirs manual CoTs for 2 experiments across the 4 benchmarks. Self-Consistency \cite{wang2023selfconsistency} is based on SP-CoT with 5 responses for each question. }
\begin{tabular}{lcccccc}
\hline
\textbf{Method}                  & \textbf{MSQ} & \textbf{HotpotQA} & \textbf{2Wiki} & \textbf{CWebQ} & \textbf{Average} \\ \hline
Zero-shot                 & 3.1/7.3   & 22.4/30.0 & 18.7/21.7 & 31.6/37.5 & 19.0/24.1 \\
Zero-shot-CoT \cite{kojima2023large}     & 5.0/8.8   & 22.6/29.6 & 24.3/27.1 & 30.3/36.2 & 20.6/25.4 \\
Manual-CoT \cite{wei2023chainofthought} \text{(Random)}      & 12.3/19.2          & 32.4/43.7         & 27.7/34.6         & 36.6/43.0      & 27.3/35.1     \\
Manual-CoT \cite{wei2023chainofthought} \text{(Cherry-Pick)} & 13.7/21.9          & 33.9/44.7         & 31.5/38.6         & 37.2/44.0      & 29.1/37.3     \\
Auto-CoT \cite{zhang2022automatic}         & 8.1/13.6  & 26.1/36.3 & 26.2/30.2 & 29.9/38.3 & 22.6/29.6 \\
SP-CoT (Ours)            & 14.5/22.6 & 33.2/42.9 & 30.1/34.7 & 37.5/43.6 & 28.8/36.0 \\
SP-CoT + Self-Consistency \cite{wang2023selfconsistency} & 18.3/28.3 & -/-       & -/-       & 47.1/54.0 & -/-      \\ \hline

\end{tabular}
\label{tab:cot}
\end{table*}

\begin{table*}[htbp]
\centering
\caption{The performance (EM) of CoT methods with recent popular LLMs on 1k subsets of test sets. We use \texttt{gpt-3.5-turbo-0301} for ChatGPT and \texttt{text-davinci-003} for InstructGPT. On smaller-scale (13B) LLMs, CoT methods achieve comparable performance boost on four MHQA benchmarks.}
\label{tab:general_test2}
\small
\begin{tabular}{lccccc}
\hline
\textbf{Method} & \textbf{ChatGPT} & \textbf{InstructGPT} & \textbf{Alpaca-13B} & \textbf{Vicuna-13B} & \textbf{Wizard-13B} \\ \hline
Zero-shot                         & 20.8 & 22.9 & 8.9  & 12.6 & 10.9 \\
Manual-CoT \cite{wei2023chainofthought} (Cherry-Pick) & 28.7 & 33.1 & 18.0 & 23.7 & 24.7 \\
Manual-CoT \cite{wei2023chainofthought} (Random)      & 26.3 & 32.4 & 17.7 & 22.0 & 24.0 \\
Auto-CoT \cite{zhang2022automatic}                & 21.2 & 31.4 & 16.9 & 21.3 & 23.3 \\
SP-CoT (Ours)                    & 26.1 & 33.1 & 17.4 & 21.5 & 22.5 \\\hline
\end{tabular}
\end{table*}

\section{Experiment Settings}
\label{app:exp}
\subsection{Hyperparameters}

This section is the experiment settings on ChatGPT (\texttt{gpt-3.5-turbo-0301}) only, for more settings of other LLMs used in our experiments, please see our code. Our code and datasets are publicly available at \href{https://github.com/noewangjy/SP-CoT}{https://github.com/noewangjy/SP-CoT}.

\subsubsection{System message}

\textit{"You should use your knowledge to answer the question to the best of your ability, not refuse to answer, even though you know your knowledge is sometimes out of date. If some references are uncertain, answer all possible cases rather than requesting further information."}

\subsubsection{Temperature}

In most cases, the default temperature is set to 0 for obtaining a most deterministic response. When we ask ChatGPT to name some terms or to generation a question, the temperature is set to 1.0 for more diversity.

\subsection{LLMs}

The 13B LLMs used in our experiments are from Huggingface Hub, use \texttt{chavinlo/gpt4-x-alpaca} for Alpaca-13B, \texttt{TheBloke/wizard-vicuna-13B-HF} for Vicuna-13B and \texttt{TheBloke/wizardLM-13B-1.0-fp16} for WizardLM-13B.

\section{Additional Experiments}

\subsection{Comparison of CoT Variants}
To provide a more comprehensive picture of current CoT methods on ODQA, we report hereby (Table \ref{tab:cot}) the performance of additional CoT variants, including Manual-CoT \cite{wei2023chainofthought} and Auto-CoT \cite{zhang2022automatic} on ChatGPT (\texttt{gpt-3.5-turbo-0301}).  

In our experiment, Manual-CoT (Cherry-Pick) adopts 8 cherry-picked questions and their CoTs manually writen by the authors. The results of Manual-CoT (Random) report the mean EM scores of randomly selected questions and theirs manual CoTs for 2 experiments across the 4 benchmarks. Self-Consistency \cite{wang2023selfconsistency} is based on SP-CoT with 5 responses for each question.

To the best of our knowledge, Self-Consistency \cite{wang2023selfconsistency} is orthogonal to existing CoT methods, including SP-CoT.  Although Self-Consistency boosts the performance of SP-CoT to a higher level (10\%-30\% increase for 5 runs), it’s worth noting that the cost of Self-Consistency is also 5 times higher.

In Table \ref{tab:general_test2}, we report the performance (EM) of CoT methods with recent popular LLMs on 1k subsets of the test sets. The scores are the average EM scores on 4 ODMR benchmarks. Although Manual-CoT \cite{wei2023chainofthought} outperforms Automated methods, it requires high quality human-labeled CoTs, which is not always accessible in real world applications. Since the cherry-picked CoTs  take the dataset features in to consideration, we consider their results as the \textbf{theoretical  upper limit} of automated approaches. Compared to previously automatic SOTA method (Auto-CoT), our proposed SP-CoT shows a decent performance boost in most cases.

\begin{table*}[htbp]
\centering
\small
\caption{The scale and average hops of each ODMR dataset generated in this paper. Please note that the scale and the average hops are largely decided by the self-generation setting and the duplication control process.}
\begin{tabular}{lrrrr}
\hline
\textbf{ } & \textbf{ChatGPT} & \textbf{Alpaca-13B} & \textbf{Vicuna-13B} & \textbf{WizardLM-13B} \\ \hline
Number of Samples       & 3550  & 1354  & 1562  & 1604 \\
Avg. Number of Hops            & 3.04  & 2.82  & 2.92  & 3.16 \\
Avg. Question tokens & 34.61 & 25.35 & 31.05 & 34.30 \\
Avg. Answer tokens   & 1.86  & 1.99  & 2.02  & 1.91 \\
\hline
\end{tabular}
\label{tab:scale}
\end{table*}

\subsection{SP-CoT on GrailQA}

We report our experiment of CoT methods on 1k subset of test set provided by GrailQA \cite{Gu_2021}. According to our ODMR setting, no external knowledge is provided to LLMs. From the results below, we notice that our proposed SP-CoT is effective on GrailQA, our results on InstructGPT (\texttt{text-davinci-003}) are presented in Table \ref{tab:grail}.

\begin{table}[htbp]
\centering
\small
\caption{Performance of CoT methods on 1k subset of test set provided by GrailQA \cite{Gu_2021}.}
\begin{tabular}{lcc}
\hline
\textbf{Method}                             & \textbf{EM}   & \textbf{F1}   \\
\hline
Zero-shot                           & 12.9 & 24.3 \\
Zero-shot-CoT \cite{kojima2023large}              & 13.5 & 25.2 \\
Manual-CoT \cite{wei2023chainofthought} (Random)        & 17.7 & 30.7 \\
SP-CoT (Ours)                      & 16.0 & 28.2 \\ \hline

\end{tabular}
\label{tab:grail}
\end{table}

\section{Constructed ODMR Datasets}

\subsection{Overview and scale}

\begin{figure*}[htbp]
    \centering
    \includegraphics[scale=1]{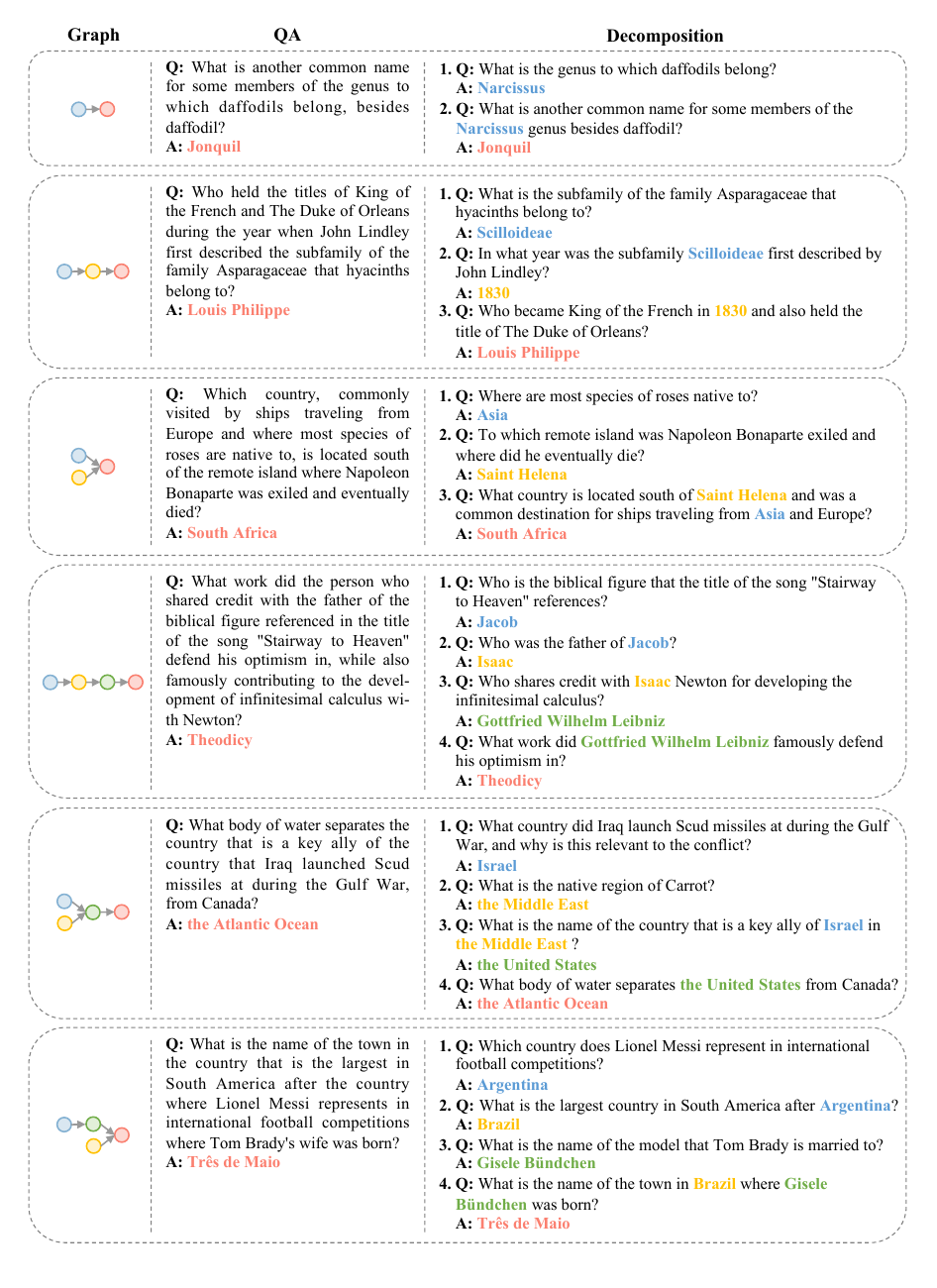}
    \caption{The illustration of six reasoning types in our automated dataset construction pipeline. These selected examples are self-generated by ChatGPT (\texttt{gpt-3.5-turbo-0301}), Vicuna-13B and WizardLM-13B.}
    \label{fig:mhqa}
\end{figure*}

For better understanding of the constructed ODMR datasets, we offer a well-designed figure (Figure \ref{fig:mhqa}) to illustrate the six types of generated questions and their step-by-step decomposition. The scale of the generated ODMR datasets is about 1-4k (Table \ref{tab:scale}), however, it’s largely dependent by the self-generation setting (how many examples to generate?) and the Duplication Control process in Stage 2 Step 2 (How many examples to keep?) To be more specific, the number of topic terms for self-generation decides the scale of generated 2-hop question pairs, and the level of duplication (how many existing question hops are allowed when constructing a new reasoning chain) decides the scale of the remaining examples after filtering.


\subsection{Topics}

The 29 manually designed topics for generation are: \textit{politicians, athletes, sports teams, sports events, countries, cities, historical figures, historical events, wars, religions, singers, songs, actors or actresses, movies or TV series, writers, books, painters, paintings, composers, classical music, tourist attractions, scientists, scientific terms, video games, animals, plants, foods, enterprises, international organizations.}

\subsection{Quality Control}

To ensure the self-generation quality, two mechanisms are included in our proposed method. 

\textbf{Self-validation in self-generation}: To ensure the quality of generated QA pairs, we employ a double-check process, where we demand the LLM to answer the generated question given the generated context and double-check If the generated answer is accordant with the target answer.

\textbf{Composability criteria in composition}: Two single-hop QA pairs $(q_1, a_1)$ and $(q_2, a_2)$ are composable into a multi-hop question $Q$ with $a_2$ as a valid answer if $a_1$ is a named entity and it is mentioned in $q_2$. Such criteria are already satisfied when our 2-hop QA pairs are generated, we use it for connecting more questions.

\subsection{Composition rather than Generation}

Directly generating k-hop questions will produce many highly-duplicated reasoning chains, which is less effective than conposition with 2-hop QAs. 

Take a connected 3-hop QAs as example: $(Q_1, A_1) \to (Q_2, A_2) \to (Q_3, A_3)$, where $A_1$ in $Q_2$, $A_2$ in $Q_3$. 

Suppose there are 2 valid next-hop QAs $(Q_4, A_4)$ and $(Q_5, A_5)$ for $(Q_3, A_3)$. Now we have 2 generated 4-hop reasoning chains: $(Q_1, A_1)\to(Q_2, A_2)\to(Q_3, A_3)\to(Q_4, A_4)$ and $(Q_1, A_1)\to(Q_2, A_2)\to(Q_3, A_3)\to(Q_5, A_5)$

which are \textbf{highly-duplicated}. When directly generating k-hop reasoning chains, the number of highly-duplicated chains will increase exponentially and such chains will be filtered out in the Duplication Control Process (Stage 2, Step 2).

Furthermore, when there are more that 2 question hops in one reasoning chain, more effort should be made to ensure direct acyclic graphs (DAGs). An example of cyclic reasoning chain is $(Q_1, A_1) \to (Q_2, A_2) \to (Q_1, A_1)$, which should be avoided.

\end{document}